\documentclass[11pt]{article}

\usepackage[final]{acl}
\usepackage{float}

\usepackage{times}
\usepackage{latexsym}

\usepackage[T1]{fontenc}

\usepackage[utf8]{inputenc}

\usepackage{microtype}

\usepackage{inconsolata}

\usepackage{graphicx}

\usepackage{amsmath}

\title{Can GPT-4o mini and Gemini 2.0 Flash Predict Fine-Grained Fashion Product Attributes? A Zero-Shot Analysis}

\author{Shubbham Shukla \\
  \texttt{ss3469@cornell.edu} \\\And
  Kunal Sonalkar \\
  \texttt{kunal.sonalkar@gmail.com} \\}

\begin{document}
\maketitle
\begin{abstract}
The fashion retail business is centered around the capacity to comprehend products. Product attribution helps in comprehending products depending on the business process. Quality attribution improves the customer experience as they navigate through millions of products offered by a retail website. It leads to well-organized product catalogs. In the end, product attribution directly impacts the `discovery experience' of the customer. Although large language models (LLMs) have shown remarkable capabilities in understanding multimodal data, their performance on fine-grained fashion attribute recognition remains underexplored. This paper presents a zero-shot evaluation of state-of-the-art LLMs that balance performance with speed and cost efficiency, mainly GPT-4o-mini and Gemini 2.0 Flash. We have used the dataset \textbf{DeepFashion-MultiModal}\footnote{\url{https://github.com/yumingj/DeepFashion-MultiModal}} to evaluate these models in the attribution tasks of fashion products. Our study evaluates these models across 18 categories of fashion attributes, offering insight into where these models excel. We only use images as the sole input for product information to create a constrained environment. Our analysis shows that Gemini 2.0 Flash demonstrates the strongest overall performance with a \textbf{ macro F1 score of 56.79\%} across all attributes, while GPT-4o-mini scored a \textbf{ macro F1 score of 43.28\%}. Through detailed error analysis, our findings provide practical insights for deploying these LLMs in production ecommerce product attribution related tasks and highlight the need for domain-specific fine-tuning approaches. This work also lays the groundwork for future research in fashion AI and multimodal attribute extraction.

\end{abstract}

\section{Introduction}

Product attribution is a foundational process in online fashion retail, underpinning the ability of ecommerce platforms to efficiently onboard sellers and manage vast, ever-growing product inventories. As fashion marketplaces scale to millions of SKUs, each with unique visual and textual characteristics, the challenge of accurately and consistently assigning detailed product attributes-such as sleeve length, fabric type, color pattern, and garment shape, becomes both critical and complex.

The scale of this problem is immense: leading fashion ecommerce platforms must process and catalog millions of new product listings from thousands of sellers, often within tight timeframes. Accurate product attribution is not only essential for inventory management and catalog organization, but also directly impacts the customer experience. Well-attributed products enable advanced filtering, faceted search, and personalized recommendations, all of which are vital for product discovery and conversion in a highly competitive market.

In practice, product attribution remains a largely manual process at various stages of the business workflow. Human annotators or seller-provided metadata are relied upon to classify and tag products, but this approach is labor-intensive, error-prone, and difficult to scale as inventory grows. Inefficiencies in this process can lead to poor catalog quality, missed sales opportunities, and suboptimal customer experiences.

To address these challenges, industry experts often wonder if large language models (LLMs) with multimodal capabilities could be leveraged. Such systems could accelerate seller onboarding, enrich product catalogs, and ultimately enhance the end-customer experience by powering more accurate search and recommendation engines. However, deploying LLMs for fine-grained product attribution at scale introduces new questions around performance, reliability, and operational cost, especially as businesses seek solutions that balance accuracy with speed and cost efficiency.

This paper makes the following contributions:
\begin{itemize}
    \item We demonstrate that in a zero-shot, image-only setting, Gemini 2.0 Flash significantly outperforms GPT-4o-Mini for fine-grained fashion attribute extraction, achieving a macro F1-score of 56.79\% to GPT-4o-Mini's 43.28\%.
    \item We establish that there is no accuracy-cost trade-off for these models; the superior performing Gemini 2.0 Flash is also more efficient, proving to be approximately 12.5\% cheaper and 24\% faster for the task.
    \item We show that deterministic model settings (low temperature) are crucial for this classification task, improving the F1-score of both models by 6-7 percentage points compared to more creative configurations.
    \item We provide a detailed attribute-level analysis that identifies specific areas of success and failure, revealing that both models handle prominent attributes like "Hat" well (F1 > 60\%) but struggle with subtle details like "Neckline" and "Waist Accessories", highlighting clear needs for domain-specific fine-tuning.
\end{itemize}

By understanding the capabilities and limitations of modern LLMs in this context, industry practitioners can make informed decisions about integrating AI-driven attribution into their workflows, enabling faster seller onboarding, richer product catalogs, and superior customer experiences, all while managing operational costs.

\begin{figure*}
    \centering
    \makebox[\textwidth]{\includegraphics[width=\textwidth]{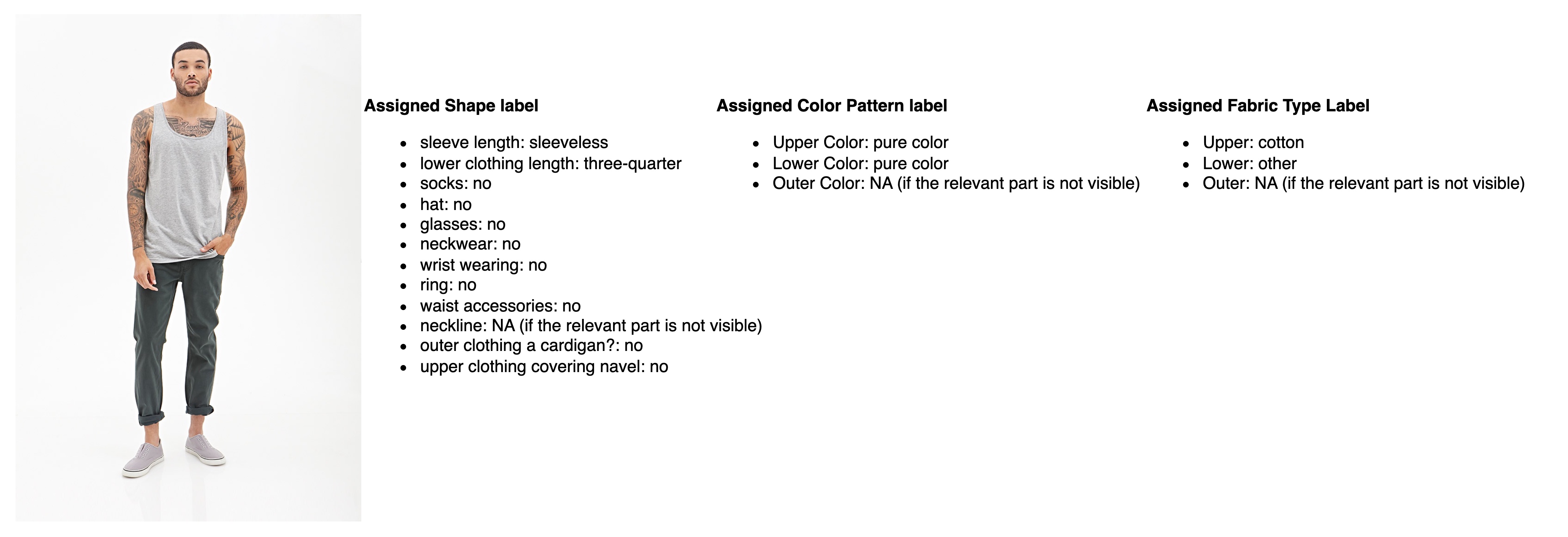}}
    \caption{Example data point}
    \label{fig:example-datapoint}
\end{figure*}

Figure \ref{fig:example-datapoint} shows an example data point from the \textbf{DeepFashion MultiModal} dataset. There are in total 18 fashion categories, 12 within shape, 3 within color pattern, and 3 within fabric type. Each of these categories have been annotated by humans. Further details about the dataset will be discussed in methodology section.

\section{Literature Review}

The task of predicting product attributes is a cornerstone of e-commerce, directly impacting catalog quality and customer-facing features like search and recommendations. Traditionally, this has been approached using deep learning models trained on large, labeled datasets. Early and even more recent sophisticated systems often rely on separate models for different modalities. For instance, a multi-modal study on the Rakuten marketplace \citep{delacomble2022} developed distinct, highly-tuned models for text (using BERT) and images (using DenseNet-121) before combining them. That work highlights the complexity of multi-modal systems, introducing a "modality-attention merger" and KL-regularization just to prevent the model from ignoring one modality over the other.

The emergence of Large Language Models (LLMs) have introduced a new paradigm. Research has rapidly shifted to evaluating the zero-shot and few-shot capabilities of these foundation models, which promises to reduce the dependency on extensive labeled datasets. Initial studies focusing on the text modality have demonstrated this potential. For instance, a detailed case study on job-type classification found that a well-prompted, zero-shot GPT-3.5-Turbo outperformed even fine-tuned supervised models like DEBERTA, highlighting that meticulous prompt engineering is a critical factor for unlocking performance \citep{clavié2023}.

Bridging the gap from purely textual analysis to visual understanding, another stream of research work has focused on adapting foundational vision-language models. A notable example is Fashion-CLIP, a version of CLIP fine-tuned on a large dataset of fashion products \citep{chia2023}. This domain-specific model demonstrated a significant performance boost over the original CLIP on fashion-related tasks like zero-shot classification and retrieval. However, this approach also inherits biases from its training data highlighting the ongoing challenge of creating truly general models. There are different prompt based frameworks like Tree of Thoughts (ToT) that are pushing the boundaries of complex reasoning in LLMs \citep{yao2023}, but their application in specialized, multimodal domains like fashion is not yet established.

Our work builds upon these foundations but carves a distinct and practical niche. While prior work has evaluated LLMs on text, built complex fine-tuned systems, or adapted models like Fashion-CLIP for specific domains, the zero-shot ability of the most recent, cost-efficient LLMs to extract fine-grained visual attributes from a single product image remains underexplored. Our paper bridges this gap by directly evaluating the practical performance of models like GPT-4o-mini and Gemini 2.0 Flash on 18 different fine-grained fashion attributes in a strictly zero-shot, image-only context. We focus on these lightweight models to assess their viability for scalable, real-world deployment where speed and cost are as important as accuracy.

\section{Preliminaries}

To establish a clear foundation for our methodology and experimental design, we define key terms and components that are central to our evaluation framework:

\textbf{Token}: In the context of Large Language Models (LLMs), a token represents the basic unit of text processing. Tokens can be words, subwords, or characters that the model uses to understand and generate text. For vision-language models, tokens also represent encoded visual features from images. Token count directly impacts API costs and model performance, making it a critical consideration for large-scale evaluations \citep{tokens}.

\textbf{Prompt}: A carefully crafted instruction or query provided to a language model to elicit specific types of responses \citep{prompting}. In our study, prompts serve as the interface between the vision-language models and the fashion attribution task. 

\textbf{GPT-4o mini}: A cost-efficient variant of OpenAI's GPT-4o model \citep{GPT4o_mini}, designed to balance performance with operational costs. GPT-4o mini maintains strong multimodal capabilities while offering significantly reduced inference costs compared to the full GPT-4o model. This makes it particularly suitable for large-scale production deployments where cost efficiency is paramount. In our evaluation, GPT-4o mini serves as a key baseline for assessing whether lightweight models can achieve competitive performance on fine-grained attribute extraction tasks.

\textbf{OpenRouter}: A unified API platform that provides access to multiple large language models through a single interface. OpenRouter \citep{OpenRouter_AI} aggregates various model providers (OpenAI, Anthropic, Google, etc.) and allows developers to switch between different models without changing their integration code. In our research, OpenRouter serves as the primary interface for accessing closed-source vision-language models, enabling consistent evaluation across different model families while providing cost tracking and rate limit management.

\textbf{Zero-Shot Prompting}: A technique where models perform tasks using only natural language instructions without any task-specific training examples or fine-tuning \citep{PromptingGuide_ZeroShot}. In our study, multimodal LLMs analyze fashion images to extract attributes using prompts that describe the task format and expected outputs, without seeing any labeled examples beforehand.

\textbf{Vision-Language Model (VLM)}: A class of neural networks capable of processing and understanding both visual and textual information simultaneously. These models can analyze images and respond to text-based queries about the visual content, making them particularly suitable for tasks like fashion product attribution where both visual features and textual descriptions are important.

\section{Architecture}

\begin{figure}[t]
  \includegraphics[width=\columnwidth]{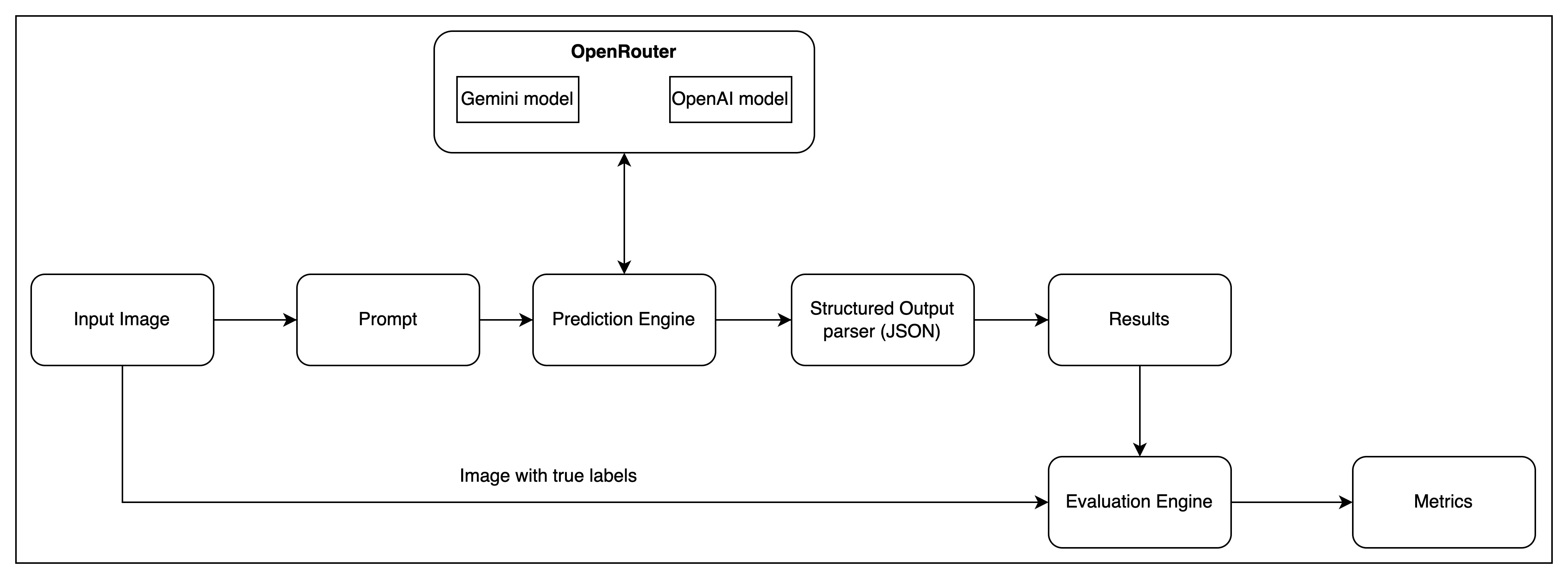}
  \caption{High Level Architecture}
  \label{fig:arch_diagram}
\end{figure}

Figure \ref{fig:arch_diagram} shows high level architecture. It's designed to enable a systematic and reproducible assessment of vision-language models for fine-grained attribute extraction. The entire process, as illustrated in the provided diagram, is orchestrated to ensure a consistent flow from data input to metric calculation.

The workflow begins with an Input Image from the DeepFashion-MultiModal dataset \citep{text2human}. This image serves as the sole input for the system, deliberately constraining the environment to test the models' pure visual understanding capabilities. The image is then passed to a Prompt generation module, which formulates a query instructing the model on the task and the desired output format.

This prompt, combined with the input image, is sent to the Prediction Engine. The engine interfaces with OpenRouter, a unified API gateway, to call the selected vision-language models-specifically Gemini 2.0 Flash or OpenAI's GPT-4o-mini. OpenRouter standardizes access to these models, simplifying the process of sending requests and receiving responses.

Once the model generates a prediction, the raw output is channeled into a Structured Output Parser. This component is critical for ensuring data integrity, as its primary function is to parse the model's response into a standardized JSON format. This step handles potential inconsistencies in the model's output and extracts the predictions into a clean, machine-readable structure, which is then stored as the Results.

Finally, the Evaluation Engine compares the structured results from the model against the ground-truth labels associated with the original input image. This engine calculates a comprehensive set of Metrics, including precision, recall, and F1-score for each of the 18 attribute categories, allowing for a detailed quantitative analysis of each model's performance.

\section{Methodology}
\subsection{Attribute Definition}

To conduct a robust evaluation of fine-grained attribute prediction, we utilize the DeepFashion-MultiModal dataset. This dataset provides a rich set of high-quality annotations ideal for our study. This dataset features manually annotated labels for clothing shapes, textures, and fabrics. The use of human-annotated data ensures a high-quality benchmark for assessing model performance.

These attributes are considered "fine-grained" because they require the model to move beyond simple object identification (e.g., recognizing a "shirt") and discern specific, detailed characteristics from the image. For example, instead of merely identifying a garment, the model must accurately classify its specific neckline style or the length of upper and lower clothes. Similarly, it must differentiate between fabric types like denim and cotton or visual patterns such as floral versus stripes. This level of detail necessitates a nuanced visual understanding that rigorously tests the descriptive power of modern vision-language models.

Our evaluation focuses on 18 distinct attribute categories, which are grouped into three primary classes as defined by the dataset's authors:

\begin{itemize}
    \item \textbf{Shape Attributes}: This is the most extensive class, detailing the structural characteristics and cut of the garments. These human-labeled attributes include properties such as the length of upper and lower clothes, specific neckline styles, and the presence of accessories.
    \item \textbf{Fabric Type Attributes}: This class specifies the material composition of the apparel. The manual annotations cover a range of common fabrics found in fashion, including denim, cotton, leather, and furry textures, among others.
    \item \textbf{Color Pattern Attributes}: This class describes the visual texture and color scheme of the clothing. The labels, annotated by humans, include categories such as solid color, stripes, floral, and other patterns.
\end{itemize}

Across all categories, the label 'NA' is used to denote instances where a relevant clothing part or attribute is not visible in the image. This allows the model to correctly handle out-of-frame items.

\subsection{Data Preparation}

The full DeepFashion-MultiModal dataset, while comprehensive, contains over 11,000 images. Evaluating models on this entire dataset would be computationally expensive and time-consuming, especially when using API-based services. To create a manageable yet representative test bed for our zero-shot evaluation, we curated a smaller subset of 1000 data points.

To ensure this subset accurately reflects the diversity of the original dataset, we employed a stratified sampling technique. The stratification was performed based on the primary product type. This approach prevents sampling bias by ensuring that our evaluation set contains a balanced distribution of different garment categories, rather than being skewed towards more common items like T-shirts or pants. 

\subsection{Using GPT-4o-mini and Gemini 2.0 Flash for Attribute Extraction}

We frame the task of fine-grained attribute extraction as a series of independent multiclass classification problems. This is achieved through a creating a prompt that instructs the LLMs to analyze an input image and classify it against the 18 predefined attribute categories.

As detailed in our prompt design (see Table \ref{table-8}), the models are explicitly asked to output numerical labels corresponding to the classes for each attribute. The task is broken down into three distinct classification challenges:

\begin{itemize}
    \item \textbf{Shape Attributes}: The models must perform 12 separate multiclass classifications. For each attribute, such as sleeve length or neckline, the model must select the most fitting class from a predefined list of options (e.g., for sleeve length, the options are 0: sleeveless, 1: short-sleeve, etc.).
    \item \textbf{Color Pattern Attributes}: The models must classify the color pattern for the Upper, Lower, and Outer clothing items visible in the image, choosing from a list of seven pattern types (e.g., 0: floral, 1: graphic, 3: pure color).
    \item \textbf{Fabric Type Attributes}: Similarly, the models must classify the fabric for the Upper, Lower, and Outer garments, selecting from a list of seven predefined fabric types (e.g., 0: denim, 1: cotton).
\end{itemize}

The prompt strictly enforces that the output for each of these three challenges must be a numerical array (e.g., [3, 3, 0, ...]). This structured output is crucial, as it allows for direct and automated comparison against the ground-truth integer labels from the dataset. By framing the task in this manner, we transform a complex visual analysis problem into a set of well-defined, quantitative classification tasks, enabling a precise evaluation of each model's zero-shot capabilities on every fine-grained attribute.

\section{Results}

Our empirical evaluation was conducted in two distinct experimental setups to assess the performance of GPT-4o-Mini and Gemini 2.0 Flash under different generation constraints. We measured precision, recall, and F1-score for each of the 18 attributes, and calculated the macro-average for each metric to represent the overall performance.

\subsection{Experiment 1: High-Creativity Setting}
In the first experiment, we configured the models with temperature=1 and top p=1. This setting allows for maximum creativity and diversity in the model's responses. Gemini 2.0 Flash significantly outperformed GPT-4o-Mini, achieving a macro F1-score of 49.72\%, which is over 12 percentage points higher than GPT-4o-Mini's score of 37.31\%. The overall performance is summarized in table \ref{table-3}. Detailed per-attribute results for this experiment can be found in the Appendix in table \ref{table-1} and \ref{table-2}.

\subsection{Experiment 2: Deterministic Setting}
For the second experiment, we used a more deterministic configuration with temperature=0 and top p=0.3. This setup constrains the model to generate more focused and predictable outputs. The overall performance for this experiment is shown in table \ref{table-6}. In this setting, the performance of both models improved. Gemini 2.0 Flash again demonstrated superior performance, achieving a macro F1-score of 56.79\%. GPT-4o-Mini's score also increased to 43.28\%. The performance gap between the two models remained significant. Detailed per-attribute results for this experiment are available in the Appendix in table \ref{table-4} and \ref{table-5}.

\subsection{Cost and Latency Analysis}
Beyond performance, we analyzed the practical cost and speed of using these models for a production-scale task. The cost and time required to process our dataset of 1000 images are presented in table \ref{table-7}. Gemini 2.0 Flash proved to be not only more accurate but also more efficient. It was approximately 12.5\% cheaper and 24\% faster than GPT-4o-Mini for this task, making it the more economically viable option for large-scale deployment.

\section{Observations}
Our results provide several key insights into the capabilities of modern, cost-efficient vision-language models for fine-grained attribute extraction.

\textbf{Gemini 2.0 Flash Superiority}: Across both experiments, Gemini 2.0 Flash consistently and decisively outperformed GPT-4o-Mini. In the more deterministic setting, Gemini achieved an F1-score nearly 13.5 percentage points higher than GPT-4o-Mini, highlighting its stronger baseline capability for this visual classification task.

\textbf{Impact of Deterministic Settings}: Both models performed better in the low-temperature (temperature=0) setting. GPT-4o-Mini saw a ~6 percentage point increase in its F1-score, while Gemini 2.0 Flash improved by ~7 percentage points. This suggests that for structured classification tasks like this, reducing model creativity leads to more reliable and accurate predictions.

\textbf{Performance Varies by Attribute}: The models showed significant variance in performance across different attributes. Both models performed relatively well on visually prominent and clearly defined attributes like Hat (F1-scores > 60\%) and Sleeve Length (F1-scores > 50\%). Conversely, they struggled with more nuanced or subtle attributes. For example, GPT-4o-Mini had particular difficulty with Neckline (F1 < 21\%) and Waist Accessories (F1 < 21\%). Gemini also found Waist Accessories challenging, though it performed considerably better on Neckline (F1 > 46\%). This indicates that while these models have strong general visual recognition, they lack the specialized knowledge to consistently identify subtle fashion details.

\textbf{Cost-Performance Synergy}: The cost analysis reveals a compelling business case for Gemini 2.0 Flash. It is not a trade-off between cost and quality; rather, the superior model is also the cheaper and faster one. This combination of higher accuracy, lower cost, and reduced latency makes it a clear choice for practical industry applications.

\section{Conclusion}

This paper presented a rigorous zero-shot evaluation of two leading cost-efficient vision-language models, GPT-4o-Mini and Gemini 2.0 Flash, on the task of fine-grained fashion attribute extraction from images. Our findings demonstrate that while both models are capable of performing this task without any specific training, their performance varies significantly.

Gemini 2.0 Flash emerged as the clear winner, delivering substantially higher accuracy across 18 attribute categories while also being faster and more cost-effective. Our analysis showed that a more deterministic model configuration (temperature=0) yields better results for this classification task. While the overall F1-score of 56.79\% for Gemini 2.0 Flash is promising for a zero-shot system, it also indicates that these models, in their current state, are not yet a complete replacement for fine-tuned systems or human annotators, especially for nuanced attributes.

Nonetheless, our work suggests that these lightweight LLMs can be useful in human-in-the-loop framework. For e-commerce platforms, they offer a powerful tool to reduce manual labor, accelerate seller onboarding, and enrich product catalogs, thereby directly improving the customer discovery experience.

\section{Future Work}

Our research opens several avenues for future investigation to build upon these findings and further bridge the gap between zero-shot performance and production-level requirements.

\textbf{Advanced Prompt Engineering}: We plan to experiment with more sophisticated prompt designs. This includes incorporating few-shot examples directly into the prompt to provide the model with in-context learning, as well as exploring chain-of-thought or step-by-step reasoning prompts to encourage more deliberate analysis of the visual evidence before making a classification.

\textbf{Benchmarking Against Specialised Models}: To better contextualize the performance of these zero-shot LLMs, a comparative analysis against traditional machine learning models is necessary. We intend to train baseline computer vision classifiers (e.g., ResNet) for each attribute and also compare the results against more established, fine-tuned vision-language models like Fashion-CLIP.

\textbf{Expanding Attribute and Dataset Scope}: Future work should expand the set of attributes to include more subjective and stylistic qualities, such as 'formality', 'style' (e.g., 'bohemian', 'minimalist'), or 'seasonality'. Evaluating the models on these less objective criteria would test their higher-level reasoning. Furthermore, validating these findings on other diverse fashion datasets would be crucial to assess the generalizability of our conclusions.

\section*{Limitations}

While this study provides a detailed zero-shot evaluation of two leading cost-efficient LLMs, it is important to acknowledge its limitations:

\textbf{Scope of Evaluation}: The primary limitation is the study's strict focus on a zero-shot, image-only context. The models were not fine-tuned on fashion-specific data, and their performance without access to textual metadata (like product descriptions), which is often available in real-world e-commerce settings, was not assessed. The study was deliberately constrained to test pure visual understanding.

\textbf{Dataset and Generalizability}: Our evaluation was conducted on a single dataset, DeepFashion-MultiModal. Although the dataset is well-annotated , the findings' generalizability to other fashion datasets with different image styles, product types, and annotation standards has not yet been validated. Furthermore, due to computational and cost constraints, the analysis was performed on a curated subset of 1,000 images, not the entire dataset of over 11,000 images.

\textbf{Model Selection:} The research deliberately centered on two cost-efficient models, GPT-4o-Mini and Gemini 2.0 Flash, to assess their viability for scalable deployment. The performance was not benchmarked against more powerful (and more expensive) flagship models or against specialized, fine-tuned vision-language models like Fashion-CLIP.

\textbf{Prompt Design}: The study utilized a single, structured prompt design that framed the task as a series of multiclass classification problems. More advanced techniques such as few-shot, chain-of-thought, or step-by-step reasoning prompts were not explored. Experimenting with these more sophisticated prompt designs could potentially improve model performance further.

\section*{Acknowledgments}
We would like to express our sincere gratitude to Yuming Jiang and the team that maintains the DeepFashion-MultiModal dataset, for granting us permission to use this valuable resource for our study.

\bibliography{custom}

\appendix

\section{Appendix}
\label{sec:appendix}
\begin{table*}
  \centering
  \begin{tabular}{llll}
    \hline
    \textbf{Attribute Value} & \textbf{GPT-4o-Mini Precision} & \textbf{GPT-4o-Mini Recall} & \textbf{GPT-4o-Mini F1}\\
    \hline
    Sleeve Length & 53.50\% & 54.80\% & 54.14\% \\ 
Lower Clothing Length & 54.00\% & 53.80\% & 53.90\% \\ 
Socks & 41.00\% & 37.70\% & 39.28\% \\ 
Hat & 57.00\% & 63.80\% & 60.21\% \\ 
Glasses & 35.00\% & 27.20\% & 30.61\% \\
Neckwear & 49.00\% & 50.30\% & 49.64\% \\
Wrist Wearing & 56.30\% & 52.60\% & 54.39\% \\
Ring & 38.60\% & 34.10\% & 36.21\% \\
Waist Accessories & 27.20\% & 16.20\% & 20.31\% \\
Neckline & 22.10\% & 19.90\% & 20.94\% \\ 
Outer clothing a Cardigan & 29.40\% & 31.60\% & 30.46\% \\ 
Upper clothing covering navel & 23.80\% & 13.20\% & 16.98\% \\
Upper Fabric & 23.90\% & 27.40\% & 25.53\% \\ 
Lower Fabric & 34.50\% & 26.80\% & 30.17\% \\ 
Outer Fabric & 39.30\% & 16.50\% & 23.24\% \\ 
Upper Color & 50.70\% & 55.20\% & 52.85\% \\ 
Lower Color & 36.50\% & 32.10\% & 34.16\% \\
Outer Color & 39.60\% & 37.50\% & 38.52\% \\
    \hline
  \end{tabular}
  \caption{\label{table-1}
    Precision, recall and F1 score for GPT-4o-Mini across 18 different fashion attributes with temperature value as 1 and $\text{top}_p$ value as 1}
\end{table*}

\begin{table*}[t!]
  \centering
  \begin{tabular*}{\textwidth}{@{\extracolsep{\fill}}lccc@{}}
    \hline
    \textbf{Attribute Value} & 
    \parbox[t]{2.5cm}{\centering\textbf{Gemini 2.0 Flash Precision}} &
    \parbox[t]{2.5cm}{\centering\textbf{Gemini 2.0 Flash Recall}} &
    \parbox[t]{2.5cm}{\centering\textbf{Gemini 2.0 Flash F1}} \\
    \hline
    Sleeve Length & 50.98\% & 51.84\% & 51.41\% \\
Lower Clothing Length & 37.83\% & 38.55\% & 38.19\% \\
Socks & 47.81\% & 52.74\% & 50.15\% \\
Hat & 61.11\% & 65.18\% & 63.08\% \\
Glasses & 38.24\% & 36.87\% & 37.54\% \\
Neckwear & 61.40\% & 59.58\% & 60.48\% \\
Wrist Wearing & 55.48\% & 57.99\% & 56.71\% \\
Ring & 52.51\% & 57.20\% & 54.75\% \\
Waist Accessories & 43.99\% & 39.91\% & 41.85\% \\
Neckline & 47.91\% & 46.09\% & 46.98\% \\
Outer clothing a Cardigan & 54.28\% & 53.47\% & 53.87\% \\
Upper clothing covering navel & 31.41\% & 33.10\% & 32.23\% \\
Upper Fabric & 33.71\% & 48.81\% & 39.88\% \\
Lower Fabric & 38.19\% & 35.68\% & 36.89\% \\
Outer Fabric & 63.87\% & 54.21\% & 58.64\% \\
Upper Color & 55.29\% & 61.58\% & 58.27\% \\
Lower Color & 50.98\% & 51.84\% & 51.41\% \\
Outer Color & 59.96\% & 65.55\% & 62.63\% \\
    \hline
  \end{tabular*}
  \caption{\label{table-2}
    Precision, recall and F1 score for Gemini 2.0 Flash across 18 different fashion attributes with temperature value as 1 and $\text{top}_p$ value as 1}
\end{table*}

\begin{table*}
  \centering
  \begin{tabular}{llll}
    \hline
    \textbf{Model Name} & \textbf{Overall Precision} & \textbf{Overall Recall} & \textbf{Overall F1}\\
    \hline
    GPT-4o-Mini & 39.52\% & 36.15\% & 37.31\% \\
Gemini 2.0 Flash & 49.16\% & 50.57\% & 49.72\% \\
    \hline
  \end{tabular}
  \caption{\label{table-3}
    Macro Precision, recall and F1 score for GPT-4o-Mini and Gemini 2.0 Flash across 18 different fashion attributes with temperature value as 1 and $\text{top}_p$ value as 1}
\end{table*}

\begin{table*}
  \centering
  \begin{tabular}{llll}
    \hline
    \textbf{Attribute Value} & \textbf{GPT-4o-Mini Precision} & \textbf{GPT-4o-Mini Recall} & \textbf{GPT-4o-Mini F1}\\
    \hline
Sleeve Length & 62.30\% & 61.10\% & 61.69\% \\
Lower Clothing Length & 54.80\% & 52.30\% & 53.52\% \\
Socks & 60.50\% & 38.00\% & 46.68\% \\
Hat & 70.10\% & 72.90\% & 71.47\% \\
Glasses & 51.90\% & 43.00\% & 47.03\% \\
Neckwear & 58.50\% & 55.50\% & 56.96\% \\
Wrist Wearing & 61.90\% & 59.60\% & 60.73\% \\
Ring & 40.20\% & 39.10\% & 39.64\% \\
Waist Accessories & 34.20\% & 20.60\% & 25.71\% \\
Neckline & 15.80\% & 14.40\% & 15.07\% \\
Outer clothing a Cardigan & 24.00\% & 21.60\% & 22.74\% \\
Upper clothing covering navel & 47.30\% & 47.60\% & 47.45\% \\
Upper Fabric & 31.70\% & 53.00\% & 39.67\% \\
Lower Fabric & 33.90\% & 30.50\% & 32.11\% \\
Outer Fabric & 46.40\% & 27.70\% & 34.69\% \\
Upper Color & 56.20\% & 59.50\% & 57.80\% \\
Lower Color & 36.20\% & 37.10\% & 36.64\% \\
Outer Color & 36.40\% & 24.70\% & 29.43\% \\
    \hline
  \end{tabular}
  \caption{\label{table-4}
    Precision, recall and F1 score for GPT-4o-Mini across 18 different fashion attributes with temperature value as 0 and $\text{top}_p$ value as 0.3}
\end{table*}

\begin{table*}[t!]
  \centering
  \begin{tabular*}{\textwidth}{@{\extracolsep{\fill}}lccc@{}}
    \hline
    \textbf{Attribute Value} & 
    \parbox[t]{2.5cm}{\centering\textbf{Gemini 2.0 Flash Precision}} &
    \parbox[t]{2.5cm}{\centering\textbf{Gemini 2.0 Flash Recall}} &
    \parbox[t]{2.5cm}{\centering\textbf{Gemini 2.0 Flash F1}} \\
    \hline
Sleeve Length & 58.72\% & 58.33\% & 58.52\% \\
Lower Clothing Length & 39.98\% & 52.09\% & 45.24\% \\
Socks & 60.07\% & 65.08\% & 62.47\% \\
Hat & 68.39\% & 71.49\% & 69.91\% \\
Glasses & 49.59\% & 50.59\% & 50.09\% \\
Neckwear & 64.59\% & 60.97\% & 62.73\% \\
Wrist Wearing & 67.71\% & 69.48\% & 68.58\% \\
Ring & 58.24\% & 60.20\% & 59.20\% \\
Waist Accessories & 24.71\% & 43.83\% & 31.60\% \\
Neckline & 43.36\% & 53.56\% & 47.92\% \\
Outer clothing a Cardigan & 62.79\% & 60.13\% & 61.43\% \\
Upper clothing covering navel & 76.15\% & 67.30\% & 71.45\% \\
Upper Fabric & 42.76\% & 51.92\% & 46.90\% \\
Lower Fabric & 41.06\% & 29.11\% & 34.07\% \\
Outer Fabric & 63.91\% & 62.26\% & 63.07\% \\
Upper Color & 65.91\% & 72.45\% & 69.03\% \\
Lower Color & 62.23\% & 50.50\% & 55.75\% \\
Outer Color & 57.52\% & 72.60\% & 64.19\% \\
    \hline
  \end{tabular*}
  \caption{\label{table-5}
    Precision, recall and F1 score for Gemini 2.0 Flash across 18 different fashion attributes with temperature value as 0 and $\text{top}_p$ value as 0.3}
\end{table*}

\begin{table*}
  \centering
  \begin{tabular}{llll}
    \hline
    \textbf{Model Name} & \textbf{Overall Precision} & \textbf{Overall Recall} & \textbf{Overall F1}\\
    \hline
    GPT-4o-Mini & 45.68\% & 42.12\% & 43.28\% \\
Gemini 2.0 Flash & 55.98\% & 58.44\% & 56.79\% \\
    \hline
  \end{tabular}
  \caption{\label{table-6}
    Macro Precision, recall and F1 score for GPT-4o-Mini and Gemini 2.0 Flash across 18 different fashion attributes with temperature value as 0 and $\text{top}_p$ value as 0.3}
\end{table*}

\begin{table*}[t!]
  \centering
  \begin{tabular*}{\textwidth}{@{\extracolsep{\fill}}lcccc@{}}
    \hline
    \textbf{Model Name} & 
    \parbox[t]{2.5cm}{\centering\textbf{Input Tokens Cost}} &  
    \parbox[t]{2.5cm}{\centering\textbf{Output Tokens Cost}} & 
    \parbox[t]{3cm}{\centering\textbf{Time to label 1000 imgs}} &
    \parbox[t]{2.5cm}{\centering\textbf{Cost for 1000 Images}} \\
    \hline
    GPT-4o-Mini & \$0.15/M Tokens & \$0.65/M Tokens & ~33 minutes & \$3.20  \\
Gemini 2.0 Flash & \$0.10/M Tokens & \$0.40/M Tokens & ~25 minutes & \$2.80  \\
    \hline
  \end{tabular*}
  \caption{\label{table-7}
    Cost and latency comparison between GPT-4o-Mini and Gemini 2.0 Flash}
\end{table*}

\begin{table*}[t!]
  \centering
  \renewcommand{\arraystretch}{1.3} 
  \begin{tabular}{p{0.9\textwidth}}
  
    You are an image analysis assistant. Given an image of clothing, please provide two labels in the following formats. Please analyze the provided image and return both responses in the specified formats. \\
    
    \vspace{1em} 

    Shape Label
    
    0. sleeve length: 0 sleeveless, 1 short-sleeve, 2 medium-sleeve, 3 long-sleeve, 4 not long-sleeve, 5 NA \newline
    1. lower clothing length: 0 three-point, 1 medium short, 2 three-quarter, 3 long, 4 NA \newline
    2. socks: 0 no, 1 socks, 2 leggings, 3 NA \newline
    3. hat: 0 no, 1 yes, 2 NA \newline
    4. glasses: 0 no, 1 eyeglasses, 2 sunglasses, 3 have a glasses in hand or clothes, 4 NA \newline
    5. neckwear: 0 no, 1 yes, 2 NA \newline
    6. wrist wearing: 0 no, 1 yes, 2 NA \newline
    7. ring: 0 no, 1 yes, 2 NA \newline
    8. waist accessories: 0 no, 1 belt, 2 have a clothing, 3 hidden, 4 NA \newline
    9. neckline: 0 V-shape, 1 square, 2 round, 3 standing, 4 lapel, 5 suspenders, 6 NA \newline
    10. outer clothing a cardigan?: 0 yes, 1 no, 2 NA \newline
    11. upper clothing covering navel: 0 no, 1 yes, 2 NA \newline
    Note: 'NA' means the relevant part is not visible. \newline
    
    Example: If you analyze an image where the sleeve length is long-sleeve, the lower clothing length is long, there are no socks, no hat, no glasses, no neckwear, no wrist wearing, no ring, no waist accessories, the neckline is round, the outer clothing is not a cardigan, and the upper clothing does not cover the navel, your response should be: [3, 3, 0, 0, 0, 0, 0, 0, 0, 2, 1, 0]. Adhere to the format. \\
    
    \vspace{1em}

    Color Pattern Label
    
    Identify Upper Color, Lower Color, and Outer Color of this clothing item. Upper Color, Lower Color, and Outer Color can be chosen from the following categories: \newline
    0: floral, 1: graphic, 2: striped, 3: pure color, 4: lattice, 5: other, 6: color block, 7: NA (if the relevant part is not visible) \newline
    
    Example: If you analyze an image where the upper clothing is blue (pure color), the lower clothing is black (pure color), and there is no outer clothing visible, your response should be: [3, 3, 7]. Adhere to the format. \\
    
    \vspace{1em}

    Fabric Type Label
    
    Identify the fabric type for the upper, lower, and outer clothing items. Fabric types can be chosen from the following categories: \newline
    0: denim, 1: cotton, 2: leather, 3: furry, 4: knitted, 5: chiffon, 6: other, 7: NA (if the relevant part is not visible) \newline
    Example: If you analyze an image where the upper clothing is made of cotton, the lower clothing is made of denim, and there is no outer clothing visible, your response should be: [1, 0, 7]. Adhere to the format. \\

    Please analyze the provided image and return both responses in the specified formats.
    
  \end{tabular}
  \caption{\label{table-8}The full prompt provided to the vision-language models for the zero-shot attribute extraction task. The prompt details the structure, categories, and required output format for the task.}
  \label{tab:prompt_details}
\end{table*}

\end{document}